\documentclass[11pt,a4paper]{article}
\usepackage{times,latexsym}
\usepackage{url}
\usepackage[T1]{fontenc}

\usepackage[acceptedWithA]{tacl2021v1}

\usepackage{xspace,mfirstuc,tabulary}

\newif\iftaclinstructions
\taclinstructionsfalse 
\iftaclinstructions
\renewcommand{\confidential}{}
\renewcommand{\anonsubtext}{(No author info supplied here, for consistency with
TACL-submission anonymization requirements)}
\newcommand{\instr}
\fi

\iftaclpubformat 

\else

\fi

\usepackage{comment}
\usepackage{url}
\usepackage{graphicx}
\usepackage{pifont}
\usepackage{booktabs}
\usepackage{enumitem}

\newif\ifdraft
\draftfalse

\ifdraft
  \newcommand{\prettycomment}[3]{\colorbox{#1}{\parbox{.8\linewidth}{#2: #3}}}
\else
  \newcommand{\prettycomment}[3]{}
\fi

\newcommand{\vacomment}[1]{\prettycomment{green}{VA}{\small#1}}

\newcommand{\ignore}[1]{}

\newcommand{\shortname}{\textsc{TopiOCQA}}
\newcommand{\longname}{Open-domain Conversational Question Answering \\ with Topic Switching}

\usepackage{caption}
\usepackage{subcaption}
\setlist[itemize]{leftmargin=*}

\title{TopiOCQA: \longname{}}

\author{
\textbf{Vaibhav Adlakha$^{1}$ \hspace{7mm} Shehzaad Dhuliawala$^{2}$} \\
\textbf{Kaheer Suleman$^{3}$ \hspace{7mm} Harm de Vries$^{4}$ \hspace{7mm} Siva Reddy$^{1,5}$} \vspace{2mm} \\
$^{1}$Mila, McGill University \hspace{4mm} $^{2}$ETH Z\"urich \hspace{4mm} $^{3}$Microsoft Montr\'eal\\
 $^{4}$ServiceNow Research \hspace{4mm} $^{5}$Facebook CIFAR AI Chair
\vspace{2mm} \\
\texttt{\small \{vaibhav.adlakha,siva.reddy\}@mila.quebec} \\
}

\begin{document}
\maketitle
\begin{abstract}

In a conversational question answering scenario, a questioner seeks to extract information about a topic through a series of interdependent questions and answers.
As the conversation progresses, they may switch to related topics, a phenomenon commonly observed in information-seeking search sessions.
However, current datasets for conversational question answering are limiting in two ways: 1) they do not contain topic switches; and 2) they assume the reference text for the conversation is given, i.e., the setting is not open-domain.
We introduce \shortname{} (pronounced Tapioca), an open-domain conversational dataset with topic switches based on Wikipedia.
\shortname{} contains 3,920 conversations with information-seeking questions and free-form answers.
On average, a conversation in our dataset spans 13 question-answer turns and involves four topics (documents).
\shortname{} poses a challenging test-bed for models, where efficient retrieval is required on multiple turns of the same conversation, in conjunction with
constructing valid responses using conversational history. 
We evaluate several baselines, by combining state-of-the-art document retrieval methods with neural reader models.
Our best model achieves F1 of 55.8, falling short of human performance by 14.2 points, indicating the difficulty of our dataset.
Our dataset and code is available at \url{https://mcgill-nlp.github.io/topiocqa}.
\end{abstract}

\section{Introduction}
\newcommand{\qstyle}[1]{\textbf{#1}}
\begin{figure}[ht!]
\footnotesize
\begin{tabular}{p{\columnwidth}}
\midrule
\texttt{Q$_1$}: \qstyle{when was the byzantine empire born what was it originally called?}\\
\texttt{A$_1$}: 5th century AD and was called Eastern Roman Empire, or Byzantium\\
\texttt{Topic}: \href{https://en.wikipedia.org/wiki/Byzantine_Empire}{Byzantine Empire}\\
\vspace{0em}
. . . . . 
\\
\vspace{0em}
\texttt{Q$_3$}: \qstyle{which battle or event marked the fall of this empire?}\\
\texttt{A$_3$}: A six-year-long civil war followed by attack from Sultan Mehmed's army\\
\texttt{Topic}: \href{https://en.wikipedia.org/wiki/Byzantine_Empire}{Byzantine Empire}\\
\vspace{0em}
\texttt{Q$_4$}: \qstyle{did he conquer other territories as well?}\\
\texttt{A$_4$}: Yes. Anatolia and in Southeast Europe as far west as Bosnia\\
\texttt{Topic}: \href{https://en.wikipedia.org/wiki/Mehmed_the_Conqueror}{Mehmed the Conqueror}\\
\vspace{0em}
\texttt{Q$_5$}: \qstyle{where is the first area located in present day terms?}\\
\texttt{A$_5$}: Turkey\\
\texttt{Topic}: \href{https://en.wikipedia.org/wiki/Anatolia}{Anatolia}\\
\vspace{0em}
. . . . .
\\
\vspace{0em}
\texttt{Q$_7$}: \qstyle{what is the present day capital of the country?}\\
\texttt{A$_7$}: Ankara\\
\texttt{Topic}: \href{https://en.wikipedia.org/wiki/Turkey}{Turkey}\\
\vspace{0em}
\texttt{Q$_8$}: \qstyle{can you name some of the other major cities here?}\\
\texttt{A$_8$}: Istanbul\\
\texttt{Topic}: \href{https://en.wikipedia.org/wiki/Turkey}{Turkey}\\
\vspace{0em}
\texttt{Q$_{9}$}: \qstyle{were any of these cities associated with the first empire you were discussing?}\\
\texttt{A$_{9}$}: The Ottomans made the city of Ankara the capital first of the Anatolia Eyalet and then the Angora Vilayet\\
\texttt{Topic}: \href{https://en.wikipedia.org/wiki/Ankara}{Ankara}\\
\bottomrule
\end{tabular}
\caption{A conversation from \shortname{}. Our dataset has information-seeking questions with free-form answers across multiple topics (documents). The consecutive turns from the same topic (document) have been excluded for brevity.}
\vspace{-1.5em}
\label{fig:example_conversation}
\end{figure}

\newcommand{\tick}{\ding{51}}
\newcommand{\cross}{\ding{55}}
\newcommand*{\TakeFourierOrnament}[1]{{%
\fontencoding{U}\fontfamily{futs}\selectfont\char#1}}
\newcommand*{\warning}{\TakeFourierOrnament{66}}
\newcommand{\citeqrecc}{{\cite{anantha&al21qrecc}}}
\newcommand{\citeorconvqa}{{\cite{chen&al20orcqa}}}
\newcommand{\citeconvques}{{\cite{christmann&al19convquestions}}}
\newcommand{\citecoqa}{{\cite{reddy&al19coqa}}}
\newcommand{\citequac}{{\cite{choi&al18quac}}}
\newcommand{\citenarrativeqa}{{\cite{kocisky&al18narrativeqa}}}
\newcommand{\citenq}{{\cite{kwiatkowski&al19nq}}}
\newcommand{\citesquad}{{\cite{rajpurkar&al18squad2}}}

\begin{table*}

\centering
\resizebox{\linewidth}{!}{\begin{tabular}{lccccc}
\toprule
\textbf{Dataset} & \textbf{Multi--turn} & \textbf{Open--domain}  & \textbf{Free--form answers} & \textbf{Information--seeking questions} & \textbf{Topic Switching} \\
\midrule
\shortname{} (ours)
& \tick & \tick & \tick & \tick & \tick\\ 
\midrule
QReCC \citeqrecc
& \tick & \tick & \tick & \warning & \warning \\
OR-QuAC \citeorconvqa
& \tick & \tick & \cross & \tick & \cross \\
CoQA \citecoqa
& \tick & \cross & \tick & \cross & \cross \\ 
QuAC \citequac
& \tick & \cross & \cross & \tick & \cross\\ 
NarrativeQA \citenarrativeqa
& \cross & \tick & \tick & \tick & \cross\\ 
Natural Questions \citenq
& \cross & \tick & \cross & \tick & \cross \\ 
SQuAD 2.0 \citesquad
& \cross & \cross & \cross & \cross & \cross \\ 
\bottomrule
\end{tabular}}

\caption{Comparison of \shortname{} with other QA datasets. \shortname{} incorporates topical changes, along with several best practices of previous datasets. \warning  represents that only a proportion of dataset satisfies the property.
}
\label{tab:datasets-comparison}
\end{table*}

People often engage in information-seeking conversations to discover new knowledge \cite{Walton2019}. 
In such conversations, a questioner (the seeker)  asks multiple rounds of questions to an answerer (the expert). 
As the conversation proceeds, the questioner becomes inquisitive of new but related topics based on the information provided in the answers \cite{stede&schlangen04infoseeking}. 
Such topic switching behaviour is natural in information-seeking conversations and is commonly observed when people seek information through search engines \cite{spink2002multitasking}.

According to \citeauthor{spink2002multitasking}, people switch from one to ten topics with a mean of 2.11 topic switches per search session.
For example, a person can start a search session about \textit{tennis}, and then land on \textit{Roger Federer}, and after learning a bit about him may land on his country \textit{Switzerland} and spend more time learning about other \textit{Swiss athletes}.
Thanks to tremendous progress in question answering research \cite{rogers2021qa}, we are coming close to enabling information-seeking conversations with machines (as opposed to just using keywords-based search).
In order to realize this goal further, it is crucial to construct datasets that contain information-seeking conversations with topic switching, and measure progress of conversational models on this task, the two primary contributions of this work.

In the literature, a simplified setting of information-seeking conversation known as conversational question answering (CQA) has been deeply explored~\citep{choi&al18quac, reddy&al19coqa}.
In this task, the entire conversation is based on a given reference text of a topic/entity.
While the CQA task is challenging, it still falls short of the real-world setting, where the reference text is not known beforehand (first limitation) and the conversation is not restricted to a single topic (second limitation).

\newcite{chen&al20orcqa} and \newcite{anantha&al21qrecc} have attempted to overcome the first limitation by adapting existing CQA datasets to the open-domain setting.
They do so by obtaining context-independent rewrites of the first question to make the question independent of the reference text. 
For example, if the reference text is about \emph{Augusto Pinochet} and the conversation starts with a question \emph{"Was he known for being intelligent?"}, the question is re-written to \emph{"Was Augusto Pinochet known for being intelligent?."}
However, as the entire question sequence in the conversation was collected with a given reference text of a topic, all the turns still revolve around a single topic.

In this work, we present \textbf{\shortname{}}\footnote{\shortname{} is pronounced as Tapioca.} -- \textbf{Topi}c switching in \textbf{O}pen-domain \textbf{C}onversational \textbf{Q}uestion \textbf{A}nswering, a large-scale dataset for information-seeking conversations in open-domain based on the Wikipedia corpus.
We consider each Wikipedia document to be a separate topic.
The conversations in \shortname{} start with a real information-seeking question from Natural Questions \cite{kwiatkowski&al19nq} in order to determine a seed topic (document), and then the questioner may shift to other related topics (documents) as the conversation progresses.\footnote{A portion of the training data also contains conversations where the questioner asks the first question given a seed topic.}
Throughout the conversation, the questioner is never shown the content of the documents (but only the main title and section titles) to simulate an information-seeking scenario, whereas the answerer has full access to the content along with the hyperlink structure for navigation.
In each turn, both questioner and answerer use free-form text to converse (as opposed to extractive text spans as is common for an answerer in many existing datasets).

Figure~\ref{fig:example_conversation} shows an example of a conversation from our dataset. 
The first question leads to the seed topic \textit{Byzantine Empire}, and after two turns switches to \textit{Mehmed the Conqueror} in \texttt{Q$_4$}, based on part of the answer (\texttt{A$_3$}) that contains reference to \textit{Mehmed}.
Note that the answers \texttt{A$_1$}, \texttt{A$_3$} and \texttt{A$_4$} are free-form answers that do not occur as spans in either the seed document or the follow up document.
The topic then switches to \textit{Anatolia} based on part of the previous answer (\texttt{A$_4$}).
The topics change in further turns to \textit{Turkey} and \textit{Ankara}.
Because of the conversational nature, \shortname{} contains questions rife with complex coreference phenomena, for instance, \texttt{Q$_9$} relies on entities mentioned in \texttt{A$_7$}, \texttt{A$_8$} and \texttt{Q$_1$}.

\shortname{} contains 3,920 conversations and 50,574 QA pairs, based on Wikipedia corpus of 5.9 million documents. On average, a conversation has 13 question-answer turns and involves 4 topics.
28\% of turns in our dataset require retrieving a document different from the previous turn.
To the best of our knowledge, \shortname{} is the first open-domain information-seeking CQA dataset that incorporates topical changes, along with other desirable properties (see Table \ref{tab:datasets-comparison}).

To investigate the difficulty of the \shortname{} dataset, we benchmark several strong retriever-reader neural baselines, considering both sparse and dense retrievers, as well as extractive and generative readers~\citep{karpukhin&al20dpr, izacard&al21fid}.
Inspired by previous work, we explore two ways to represent the question: (1) concatenating the entire conversation history~\cite{chen&al20orcqa}, and (2) self-contained rewrites of the conversational question~\cite{anantha&al21qrecc}.
The best performing model -- Fusion-in-Decoder~\cite{izacard&al21fid} trained on concatenated conversation history -- is 14.2 F1 points short of human performance, indicating significant room for improvement.
We also evaluate GPT-3 to estimate the performance in a closed-book zero-shot setting, and its performance is 38.2 F1 points below the human performance.

\section{Related Work}

\subsection{Open-Domain Question Answering}
In open-domain question answering, a model has to answer natural language questions by retrieving relevant documents.
This can be considered as a simplified setting of open-domain CQA, where the conversation is limited to just one turn.
Several datasets have been proposed for this task. 
On one hand, reading comprehension datasets like SQuAD \cite{rajpurkar&al16squad, rajpurkar&al18squad2} which consist of (question, document, answer) triplets, have been adapted for the task by withholding access to the document \cite{chen&al17drqa}. While these datasets have been helpful in spurring modelling advances, they suffer from an annotator bias because they were not collected in an information-seeking setup. That is, annotators had access to the target answer and its surrounding context and therefore formulated questions that had a high lexical overlap with the answer~\citep{jia&liang17adversarialsquad}.
On the other hand, web-search based datasets do not suffer from such artefacts because they are curated from real search engine queries. 
The WikiQA \cite{yang&al15wikiqa} and MS Marco \cite{nguyen&al16msmarco} datasets contain queries from the Bing search engine, whereas Natural Questions \cite{kwiatkowski&al19nq} contain queries from the Google search engine.

Models for open-domain QA often follow a two-stage process: (1) A retriever selects a small collection of documents relevant to the question from a big corpus (e.g. Wikipedia), (2) a reader extracts or generates an answer from the selected documents.
While classical approaches rely on counting-based bag-of-words representations like TF-IDF or BM25  \cite{chen&al17drqa,  wang&al18r3, yang&al19bertserini}, more recent deep learning approaches learn dense representations of the questions and document through a dual-encoder framework \cite{lee&al19orqa, karpukhin&al20dpr}.
In such learned retriever setups, document retrieval is done efficiently using Maximum Inner Product Search (MIPS, \citealt{shrivastava&li14mips}).

\subsection{Conversational Question Answering (CQA)}
CQA extends the reading comprehension task from a single turn to multiple turns. Given a reference document, a system is tasked with interactively answering a sequence of information-seeking questions about the corresponding document.
This conversational extension leads to novel challenges in modeling linguistic phenomena such as anaphora (referencing previous turns) and ellipsis (omitting words from questions), as well as in performing pragmatic reasoning.
Large-scale conversational datasets such as CoQA~\cite{reddy&al19coqa} and QuAC~\cite{choi&al18quac} 
have facilitated much of the research in this area.
These datasets differ along several dimensions, two of which are (1) CoQA has short free-form answers, whereas QuAC has long extractive span-based answers, and (2) unlike CoQA, QuAC is collected in a simulated information-seeking scenario.

Models for CQA have used simple concatenation of the question-answer history~\cite{zhu&al19sdnet}, history turn selection~\cite{qu&al19historybert, qu&al19history}, and question-rewrites~\cite{vakulenko&al21quesrewite}. For question-rewriting, a different module is trained on self-contained rewrites of context-dependent questions. For example, a plausible rewrite of \texttt{Q$_8$} (Figure \ref{fig:example_conversation}) is
\emph{``can you name some of the major cities in Turkey apart from Ankara?''}. The re-written question is then answered using open-domain QA systems.
Two popular question-rewriting datasets for training this module are 
(1) CANARD~\cite{elgohar&tal19canard}, which contains re-writes of ~50\% of QuAC, and (2) QReCC~\cite{anantha&al21qrecc}, which contains rewrites of the entire QuAC dataset and a small portion from other sources.

\subsection{Open-Domain CQA}
In this work, we focus on constructing a challenging benchmark for open-domain CQA. 
The open-domain aspect requires systems to answer questions \textit{without} access to a reference document. 
The conversational aspect enables users to ask multiple related questions, which can, in principle, span several different topics. With \shortname{}, we introduce the first open-domain CQA dataset that explicitly covers such topical switches.

Previous datasets for this task re-purpose existing CQA datasets. The OR-QuAC dataset~\cite{chen&al20orcqa} is automatically constructed from QuAC~\cite{choi&al18quac}  and CANARD~\cite{elgohar&tal19canard} by replacing the first question in QuAC with context-independent rewrites from CANARD. QReCC~\cite{anantha&al21qrecc} is a large-scale open-domain CQA and question rewriting dataset which contains conversations from QuAC, TREC CAsT~\cite{dalton&al20cast} and Natural Questions (NQ;~\citealt{kwiatkowski&al19nq}).
All the questions in OR-QuAC and 78\% of questions in QReCC are based on QuAC. 
As conversations in QuAC were collected with a given reference document, the question sequences of these conversations revolve around the topic or entity corresponding to that document.
21\% of questions in QReCC are from NQ-based conversations. 
As NQ is not a conversational dataset, the annotators of QReCC use NQ to start a conversation. 
A single annotator is tasked with providing both follow-up questions and answers for a given NQ question. 
In contrast to QReCC, conversations in our dataset are collected in a simulated information-seeking scenario using two annotators (Section \ref{sec:topic-switching-interface}).

Deep learning models for this task have followed a similar retriever-reader setup as open-domain QA. 
Instead of a single question, previous works have explored feeding the entire conversation history \cite{chen&al20orcqa}, or a context independent re-written question \cite{anantha&al21qrecc}.

\section{Dataset Collection}
\label{sec:dataset_collection}
Each conversation in \shortname{} is an interaction between two annotators -- a \textit{questioner} and an \textit{answerer}.
The details about the annotator selection are provided in Appendix~\ref{appendix_sec:data_collection}.

\subsection{Seed topics and document collection}

The seed topics essentially drive the conversation.
In order to make them interesting for annotators, we select the \textit{good}\footnote{\href{https://en.wikipedia.org/wiki/Wikipedia:Good_articles}{Wikipedia Good articles}} articles of Wikipedia as seed topics (around 35k) for the first turn, but use entire Wikipedia for later turns.
We used the Wikipedia dump from 10/20/2020, which consists of ~5.9 million documents.
We used Wikiextractor\footnote{\href{https://github.com/attardi/wikiextractor}{github:wikiextractor}} to extract the text.
While pre-processing the Wikipedia documents, we retain the hyperlinks that refer to other Wikipedia documents, thus ensuring that we can provide all the documents requested by annotators (via hyperlinks) during the conversation.

\subsection{Simulating information-seeking scenario} 
Information-seeking conversations are closer to the real-world if an information need can be simulated via the data collection interface. 
In \shortname{}, we achieve this by withholding questioner's access to the full reference text of the document.
The questioner can only see the metadata (main title and the section titles) of the Wikipedia documents,
whereas the answerer can access the entire text of the documents. 
On finding the answer, the answerer highlights a contiguous span of text as rationale, and generates a free-form answer. 
The answerer also has the option to mark the question as \textit{unanswerable}. 
The conversation history is visible to both the annotators.

As a conversation starting point, the first question is sampled from a subset of Natural Questions (NQ; \citealt{kwiatkowski&al19nq}) since NQ contains genuine information-seeking questions asked on Google.
We only sample those questions for which the answer is in our seed document pool.
To increase the diversity of our dataset, we also allow the questioner to formulate the first question based on the provided seed topic entity for 28\% of the conversations.

\subsection{Enabling topic-switching}
\label{sec:topic-switching-interface}
The key feature of the interface is enabling topic switching via hyperlinks. For the answerer, the text of the document includes clickable hyperlinks to other documents.
On clicking these links, the current document in the answerer's interface changes to the requested (clicked) document. This enables the answerer to search for answers in documents beyond the current one.
The questioner can access the metadata of documents visited by the answerer and documents present in the rationale of the answers.
For example, let us assume that given the seed document \textit{Daniel Radcliffe} and the first question \textit{``Where was Daniel Radcliffe born?''}, the answerer selects \textit{``Daniel Jacob Radcliffe was born in London on 23 July 1989''} span as rationale and provides \textit{``London''} as the answer. If \textit{London} is a hyperlink in the rationale span, then the metadata of both \textit{Daniel Radcliffe} and \textit{London} is available to the questioner to form the next question.
If the next question is \textit{``What is its population?''}, the answerer can switch the current document from \textit{Daniel Radcliffe} to \textit{London} by clicking on the hyperlink, and can then find and provide the answer.
The conversation up till this point involves two topics -- \textit{Daniel Radcliffe} and \textit{London}.
We also provide easy navigation to previously visited documents for both the annotators.
This interface design (Figure~\ref{fig:interface-screenshot}) ensures that information about the new topic is semantically connected to topics of the previous turns, similar to natural human-human conversations \cite{sacks&jefferson71conversation}.

\subsection{Additional annotations}
To account for multiple valid answers,
we collected three additional annotations for answers of conversations in evaluation sets (development and test splits).
For this task, at any turn, the annotator can see all the previous questions and original answers. 
Showing original answers of previous turns is important in a conversational setting as the subsequent questions can potentially depend on them.
We also provide the list of documents corresponding to previous turns of the original conversation.
This ensures that the current annotator has all the information the original answerer had while providing the answer.
Similar to the answerer, the annotator then provides the rationale and the answer, or marks the question as \textit{unanswerable}.

\begin{figure*}[ht!]
\begin{subfigure}{0.65\textwidth}
\includegraphics[width=\columnwidth]{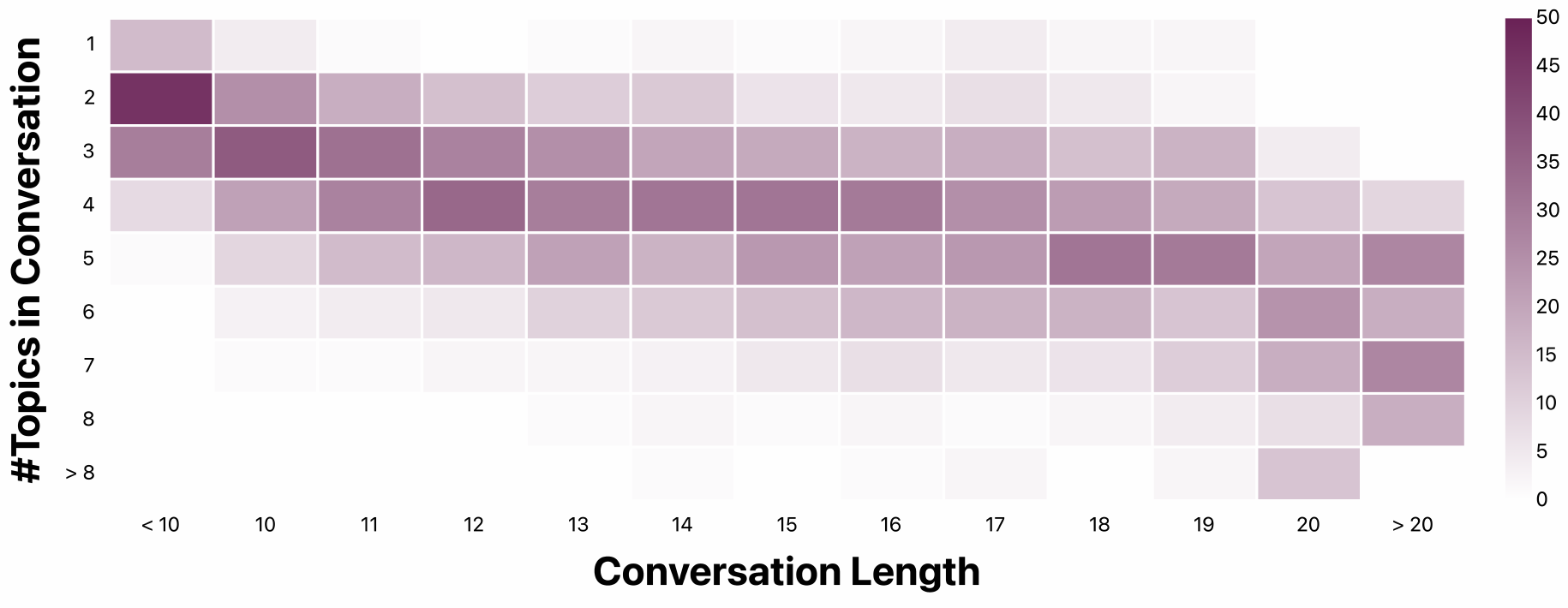}
\caption{} 
\label{fig:topic-heatmap}
\end{subfigure}
\hspace{0.5em}
\begin{subfigure}{0.35\textwidth}
\includegraphics[width=0.99\columnwidth]{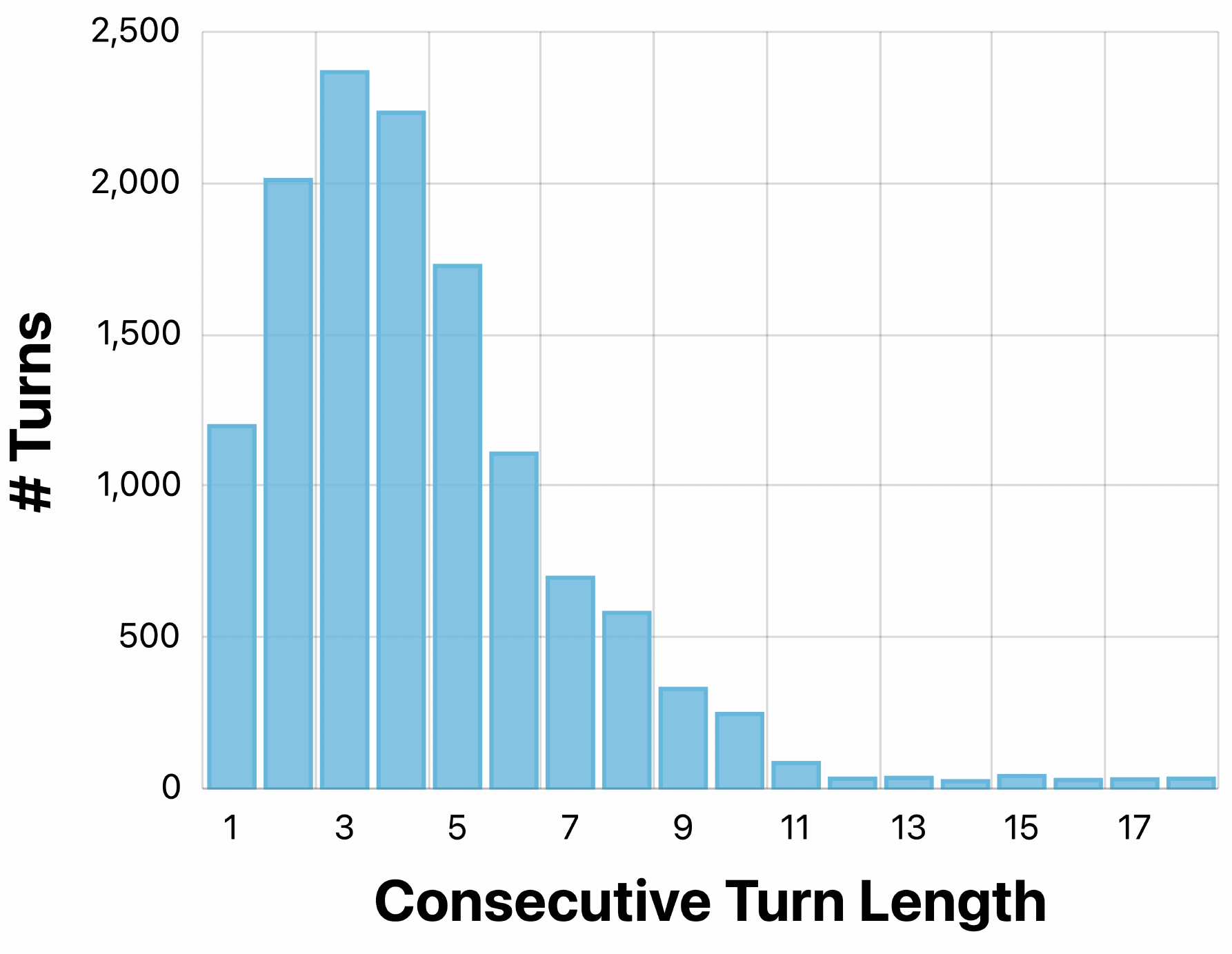}
\caption{}
\label{fig:topic-bar-chart-analysis}
\end{subfigure}
\caption{Analysis of the topic switches in \shortname{}. In (a) we show the distribution of the number of topics (in percentage) for each conversation length. Longer conversations typically include more topics. In (b) we show a histogram of the topic lengths, illustrating that usually 3-4 consecutive questions stay within the same topic.}
\label{fig:topic-switching-stats}
\end{figure*}

\section{Dataset Analysis}
\label{sec:dataset_analysis}



\begin{table}[t]
\centering
\resizebox{\linewidth}{!}{\begin{tabular}{lrrr|r}
\toprule
\textbf{Dataset} & \textbf{Train} & \textbf{Dev} & \textbf{Test} & \textbf{Overall}\\
\midrule
\# Turns
& 45,450 & 2,514 & 2,502 & 50,466\\
\# Conversations
& 3,509 & 205 & 206 & 3920\\
\# Tokens / Question
& 6.91 & 6.89 & 7.11 & 6.92\\
\# Tokens / Answer
& 11.71 & 11.96 & 12.27 & 11.75\\
\# Turns / conversation
& 13 & 12 & 12 & 13\\
\# Topics / conversation
& 4 & 4 & 4 & 4\\
\bottomrule
\end{tabular}}

\caption{Dataset statistics of \shortname{}}


\label{tab:dataset-analysis}
\end{table}


We collected a total of 3,920 conversations, consisting of 50,466 turns. The annotators were encouraged to complete a minimum of 10 turns. Conversations with fewer than 5 turns were discarded. 
We split the data into train, development and test splits.

Table~\ref{tab:dataset-analysis} reports simple statistics of the dataset splits. On average, a conversation in \shortname{} has 13 question-answer turns and is based on 4 documents. Our dataset differs from other conversational question-answering datasets by incorporating topic switches in the conversation.

\begin{figure*}[ht!]
\includegraphics[width=\linewidth]{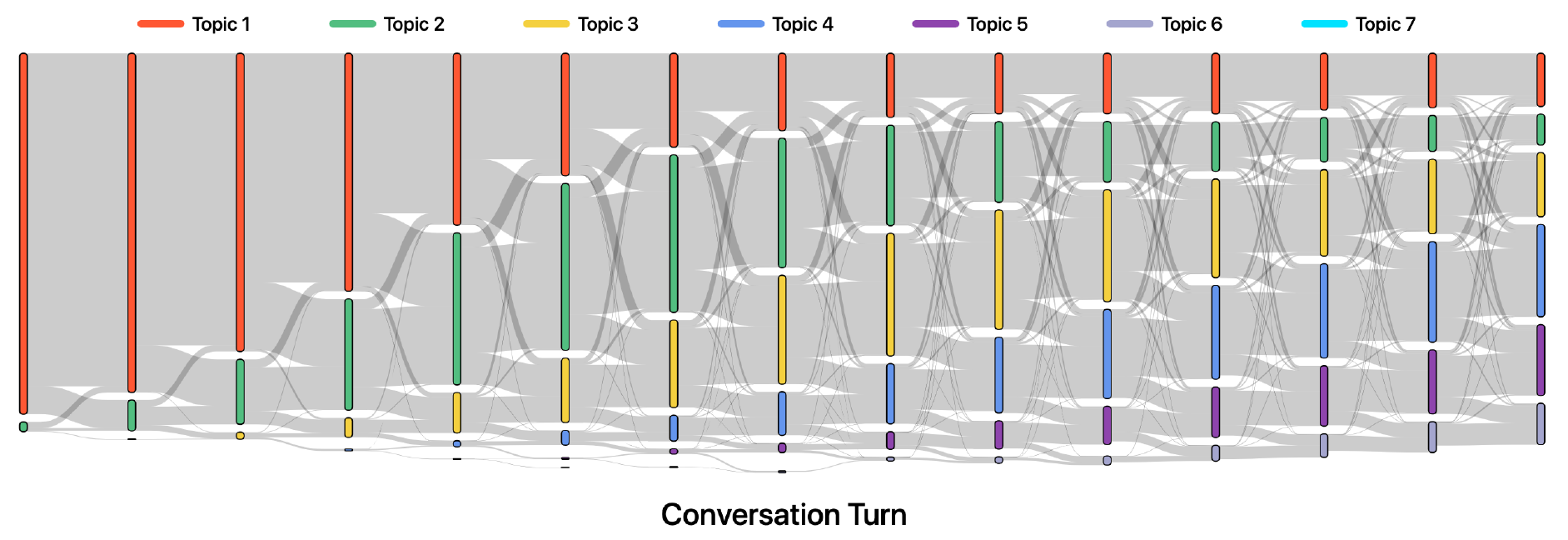}
\caption{A flow diagram of topic switches over conversations up to 15 turns. There are complex interactions between the topics, especially later in the conversation. \vacomment{want to label X axis but it is taking too long in the JS library that I am using, so skipping it for now}
\vacomment{Image commented out for faster loading}}
\label{fig:topic-switching-sankey}
\end{figure*}

\begin{table*}[t]
\begin{minipage}{0.4\textwidth}
\centering
\hspace{-1em}\includegraphics[width=0.8\columnwidth]{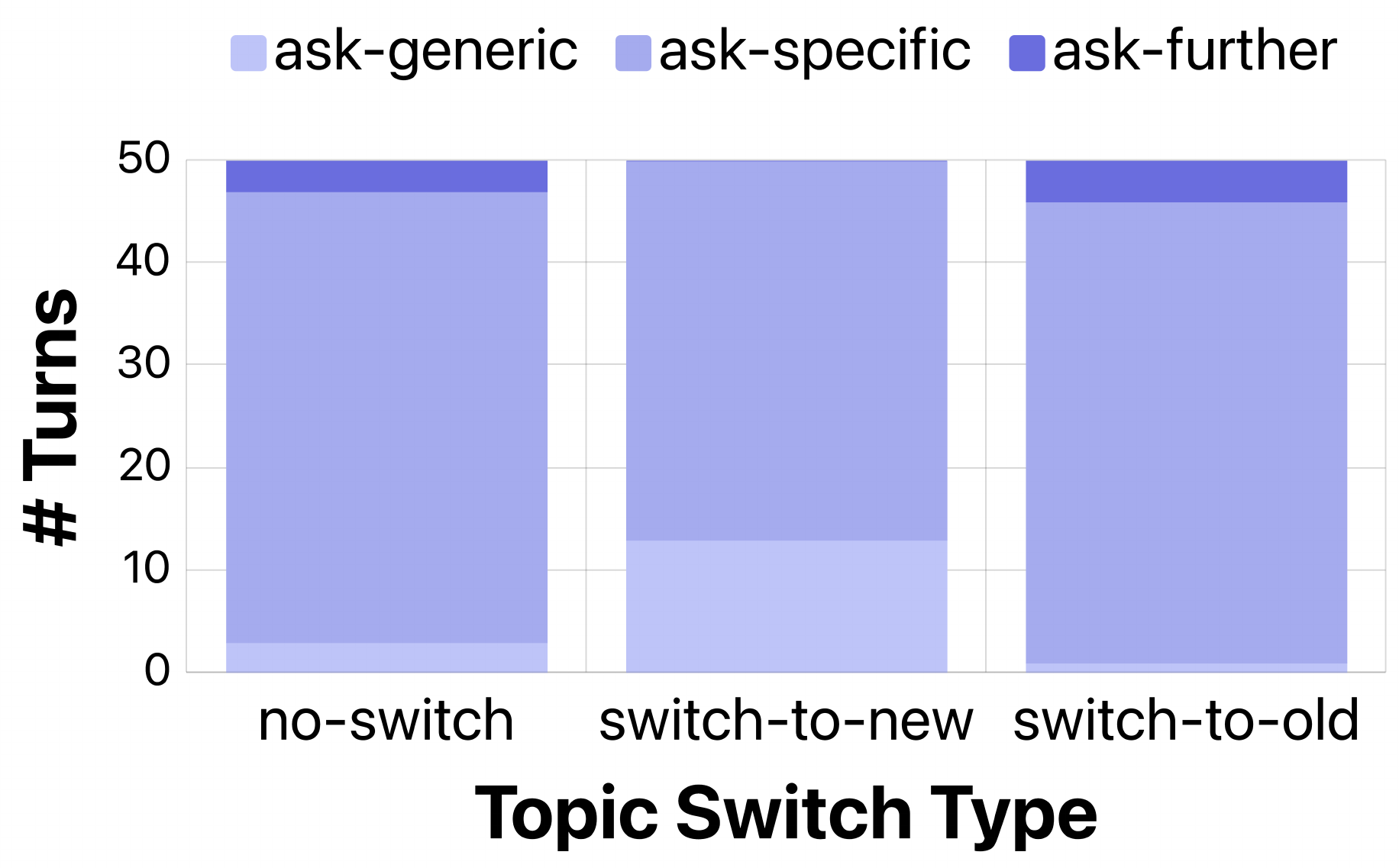}
\captionof{figure}{Distribution of various question types for each turn type. Questions asking about specific attributes are most common. Generic questions are likely to be observed when switching to a new topic.}
\label{fig:human_annotation_column}
\vspace{1em}

\centering
\resizebox{0.8\linewidth}{!}{
\begin{tabular}{l r}
\toprule
Question Type & Avg Answer length \\
\midrule
\texttt{ask-generic} & 22.43 \\
\texttt{ask-specific} & 11.38 \\
\texttt{ask-further} & 11.23 \\
\bottomrule
\end{tabular}
}
\captionof{table}{Average answer length of different question types. Generic questions tend to have longer answers.}
\label{tab:answer-length-annotations}
\end{minipage}
\hspace{0.8em}
\begin{minipage}{0.55\textwidth}



\centering
\resizebox{\linewidth}{!}{
\begin{tabular}{l l l}
\toprule
Turn type & Question type & Conversation turn\\
\midrule
\texttt{no-switch} & \texttt{ask-generic} & Q: who is mariah carey?\\
& & A: An American singer \\ & & songwriter and actress \\
& & Topic: \href{https://en.wikipedia.org/wiki/Mariah_Carey}{Mariah Carey}\\
\\
\texttt{no-switch} & \texttt{ask-specific} & Q: name one of her famous songs.\\
& & A: Oh Santa! \\
& & Topic: \href{https://en.wikipedia.org/wiki/Mariah_Carey}{Mariah Carey}\\
\\
\texttt{switch-to-new} & \texttt{ask-specific} & Q: how was it received?\\
& & A: There were mixed reviews \\
& & Topic: \href{https://en.wikipedia.org/wiki/Oh_Santa!}{Oh Santa!} \\
\\
\texttt{switch-to-old} & \texttt{ask-specific} & Q: is she married?\\
& & A: Yes \\
& & Topic: \href{https://en.wikipedia.org/wiki/Mariah_Carey}{Mariah Carey}\\
\\
\texttt{no-switch} & \texttt{ask-further} & Q: to whom?\\
& & A: Tommy Mottola \\
& & Topic: \href{https://en.wikipedia.org/wiki/Mariah_Carey}{Mariah Carey}\\
\bottomrule
\end{tabular}
}

\caption{Example of a conversation various turn types and question types. Random samples of each turn type are manually annotated with one of the question types.}
\label{tab:human-annotation}
\end{minipage}
\end{table*}
\subsection{Topic Switching}\label{sec:topicswitching}
Before we start our analysis, let us first define the notion of a topic switch in \shortname{}. Recall that answers are based on Wikipedia articles, where each document consists of several sections. While one can argue that a topic switch occurs when the answer is based on a different section of the \emph{same} document, we opt for a more conservative notion and define a switch of topic if the answer is based on a \emph{different} Wikipedia document.  

\paragraph{Number of topics vs conversation length} We begin our analysis by investigating how the number of topics varies with the conversation length. In Figure~\ref{fig:topic-heatmap} we show a heat-map of the number of topics for each conversation length, where each column is normalized by the number of conversations of that length. We observe that longer conversations usually include more topics. Most 10-turn conversations include 3 topics, 14-turn conversations include 4 topics, and 18-turn conversations include 5 topics. 
The conversations with fewer than 10 turns mostly include just 2 topics.

\paragraph{Topic flow in conversation} Next, we examine how often consecutive questions stay within the same topic. To do so, we first cluster conversations into sequences of turns for which all answers are from the same document. Then, we count how many turns belong to topic clusters of a particular length. Figure~\ref{fig:topic-bar-chart-analysis} shows the distribution of topic lengths. The mode of the distribution is 3, signifying that annotators usually ask 3 questions about the same topic before switching. Asking 2 or 4 consecutive questions on the same topic is also frequently observed. However, we rarely see more than 10 consecutive turns on the same topic.

We also analyse the flow of topics throughout the conversation. Do annotators always introduce new topics or do they also go back to old ones?  Figure~\ref{fig:topic-switching-sankey} depicts a flow diagram of topics in conversations up to 15 turns. Note that we have indexed topics according to its first occurrence in the conversation. We can see that the majority of switches introduce new topics, but also that more complex topic switching emerges in later turns. Specifically, we see that, from sixth turn onwards, questioners frequently go back one or two topics in the conversation. Overall, this diagram suggests that there are complex interactions among the topics in the conversation.

\paragraph{Qualitative assessment of topic switching} In order to understand the nature of a topic switch, inspired from \citet{stede&schlangen04infoseeking}, we classify questions into three types: \texttt{ask-generic} refers to general open-ended questions, \texttt{ask-specific} questions ask about a specific attribute or detail of a topic, and \texttt{ask-further} is a question type that seeks additional details of an attribute discussed in one of the previous turns. Table~\ref{tab:human-annotation} shows examples of each type for questions in the same conversation. We consider three types of turns for our evaluation. If the answer document of the turn is same as the previous turn, we refer to it as \texttt{no-switch}. If a topic switch has happened, and the answer document is present in one of the previous turns, it is considered to be \texttt{switch-to-old}. The final category, \texttt{switch-to-new} refers to turns where current answer document has not been seen in the conversation before. These different types of topic switches are also illustrated in Table~\ref{tab:human-annotation}.

We sample 50 turns of each type, and manually label them with one of the three question types. Figure~\ref{fig:human_annotation_column} shows the results of our evaluation. \texttt{ask-specific} is the most common question type across all types of turns, indicating that most of the questions in the dataset focus on specific attributes of a topic. \texttt{ask-generic} has a much higher proportion in \texttt{switch-to-new} turn types, indicating that it is more likely to see generic questions in turns that introduce a new topic in the conversation, compared to other turn types. \texttt{ask-further} has almost equal proportion in \texttt{no-switch} and \texttt{switch-to-old}, with \texttt{switch-to-old} being slightly higher. \texttt{ask-further} is not observed in \texttt{switch-to-new} as follow-up questions are generally not possible without the topic being discussed in any of the previous turns.

We also look at average answer length of answers of all three question types (Table~\ref{tab:answer-length-annotations}). Unsurprisingly, \texttt{ask-generic} has a much higher answer length compared to other types, presumably due to the open-ended nature of the question.

\section{Experimental Setup}
\label{sec:models}
The task of open-domain information-seeking conversation can be framed as follows. Given previous questions and ground truth answers $\{q_1, a_1, q_2, a_2, \ldots, q_{i-1}, a_{i-1}\}$ and current question $q_i$, the model has to provide the answer $a_i$. This can be considered as an \textit{oracle} setting, as the gold answers of previous questions are provided. The models can optionally use a corpus of documents $C = \{d_1, d_2, \ldots, d_N\}$.

\subsection{Models}
We consider models from two categories, based on whether they use the document corpus or not. 
The \textit{closed-book} models use just the question-answer pairs, whereas \textit{open-book} models use the document corpus, along with question-answer pairs. We now describe the implementation and technical details of both classes of models. 

\subsubsection{Closed-book}
Large-scale language models often capture a lot of world knowledge during unsupervised pre-training \cite{petroni&al19lama,roberts&al20t5short}. These models, in principle, can answer questions without access to any external corpus. We consider GPT-3 \cite{brown&al20gpt3} -- an autoregressive language model with 175 billion
parameters, and evaluate it on \shortname{}.
The input to GPT-3 is a prompt\footnote{\href{https://beta.openai.com/examples/default-qa}{beta.openai.com/examples/default-qa}} followed by previous question-answer pairs and the current question. Since GPT-3 is never explicitly exposed to any training examples, this can be considered as a \textit{zero-shot} setting.

\subsubsection{Open-book}
We build on state-of-the-art QA models that adapt a two step retriever-reader approach. For the retriever, we consider BM25 \cite{robertson&al95bm25} and DPR Retriever \cite{karpukhin&al20dpr}. Given a query, BM25 ranks the documents based on a bag-of-words scoring function. On the other hand, DPR learns dense vector representations of document and query, and uses the dot product between them as a ranking function.

We consider two types of neural readers - (1) DPR Reader \cite{karpukhin&al20dpr}, which re-ranks the retrieved passages and selects a span from each document independently. The span with highest span score is chosen as the answer. (2) Fusion-in-Decoder (FiD; \citealt{izacard&al21fid}), which encodes all retrieved passages independently, and then jointly attends over all of them in the decoder to generate the answer.

\begin{figure}[t!]
\footnotesize
\begin{tabular}{p{\columnwidth}}
\midrule
\texttt{Q$_1$}: who is lead singer of rage against the machine? \\
\texttt{A$_1$}: Zack de la Rocha\\
\vspace{0em}
\texttt{Q$_2$}: when was it formed?\\
\texttt{A$_2$}: 1991\\
\vspace{0em}
\texttt{Q$_3$}: was it nominated for any award?\\
\midrule
\textbf{\textsc{Original}}: was it nominated for any award\\
\vspace{0em}
\textbf{\textsc{AllHistory}}: who is lead singer of rage against the machine \texttt{[SEP]} Zack de la Rocha \texttt{[SEP]} when was it formed? \texttt{[SEP]} 1991 \texttt{[SEP]} was it nominated for any award \\
\vspace{0em}
\textbf{\textsc{Rewrites}}: was rage against the machine nominated for any award\\
\bottomrule
\end{tabular}
\caption{A partial conversation and different question representations of \texttt{Q$_3$}. The \textsc{Rewrites} representation is an example, not the output of our \textit{QR} module.}
\label{fig:example_question_representation}
\end{figure}

For these models, we consider three different question representations for question at $n^{th}$ turn of the conversation ($q_n$).
Figure~\ref{fig:example_question_representation} shows an example of different question representations for the third question (\texttt{Q$_3$}) of a conversation.

\begin{itemize}
    \item \textbf{\textsc{Original}}: This serves as a naive baseline where just the current question $q_n$ is passed to the model.
    \item \textbf{\textsc{AllHistory}}: The question is represented as $q_1$ \texttt{[SEP]} $a_1$ \texttt{[SEP]} $q_2$ \texttt{[SEP]} $a_2$ \texttt{[SEP]} $\ldots$ \texttt{[SEP]} $q_{n-1}$ \texttt{[SEP]} $a_{n-1}$ \texttt{[SEP]} $q_n$. When constrained by the encoder input sequence length, we retain the first turn and as many turns prior to the current turn as possible, i.e.
    $k$ is chosen such that $q_1$ \texttt{[SEP]} $a_1$ \texttt{[SEP]} $q_{n-k}$ \texttt{[SEP]} $a_{n-k}$ \texttt{[SEP]} $\ldots$ \texttt{[SEP]} $q_{n-1}$ \texttt{[SEP]} $a_{n-1}$ \texttt{[SEP]} $q_n$ satisfies encoder input limits.
    \item \textbf{\textsc{Rewrites}}: Given a query-rewriting module $QR$, let $q^\prime_n=QR(q_1, a_1, \ldots, q_{n-1}, a_{n-1}, q_n)$ denote the decontextualized question, conditioned on the conversation history. $q^\prime_n$ is then passed to the model.
\end{itemize}

\noindent\citet{wang&al19multiBERT} observed that fixed-length text segments from documents are more useful than full documents in both retrieval
and final QA accuracy. 
Hence, we split a Wikipedia document into multiple text blocks of at least 100 words, while preserving section and sentence boundaries. These text blocks, augmented with the metadata (main title and section title) are referred to as \textit{passages}. 
This resulted in 25.7 million passages, which act as basic units of retrieval. To form question-passage pairs for training DPR Retriever, 
we select the passage from gold answer document that contains the majority of rationale span.

Following the original works, we use BERT \cite{devlin&al19bert} for DPR (both Retriever and Reader) and T5 \cite{raffel&al20t5} for FiD as base models. Since DPR Reader requires a span from passage for each training example, we heuristically select the span from the gold passage that has the highest lexical overlap (F1 score) with the gold answer.
For the query-rewriting module $QR$, we fine-tune T5 model on rewrites of QReCC \cite{anantha&al21qrecc}, and use that to generate the rewrites for \shortname{}. 
We refer the reader to Appendix~\ref{appendix_sec:query_rewriting} for more details. The hyperparameters for all models are mentioned in Appendix~\ref{appendix_sec:hyperparam}.

\setlength{\tabcolsep}{12pt}
\begin{table*}[ht]
\small
\centering {\begin{tabular}{lcccccccc}
\toprule       
Model & Question Rep & \multicolumn{2}{c}{Dev} & \multicolumn{2}{c}{Test} \\ 
\cmidrule(r){3-4} \cmidrule(r){5-6}

& & EM & F1
& EM & F1
\\

\midrule      
Human & 
& \textbf{40.2} & \textbf{70.1}
& \textbf{40.3} & \textbf{70.0}
\\
\midrule
\midrule

GPT-3 & 
& 12.4 & 33.4
& 10.4 & 31.8
\\
\midrule

& \textsc{Original}
& 7.1 & 12.8
& 7.2 & 13.0
\\
BM25 + DPR Reader & \textsc{AllHistory}
& 13.6 & 25.0
& 13.8 & 25.2
\\
& \textsc{Rewrites}
& 15.4 & 32.5
& 15.7 & 31.7
\\
\midrule

& \textsc{Original}
& 10.1 & 21.8
& 10.5 & 22.6
\\
BM25 + FiD & \textsc{AllHistory}
& 24.1 & 37.2
& 23.4 & 36.1
\\
& \textsc{Rewrites}
& 24.0 & 41.6
& 24.9 & 41.4
\\
\midrule

& \textsc{Original}
& 4.9 & 14.9
& 4.3 & 14.9
\\
DPR Retriever + DPR Reader & \textsc{AllHistory}
& 21.0 & 43.4
& 19.4 & 41.1
\\
& \textsc{Rewrites}
& 17.2 & 36.4
& 16.5 & 35.2
\\
\midrule

& \textsc{Original}
& 7.9 & 21.6
& 7.8 & 21.4
\\
DPR Retriever + FiD & \textsc{AllHistory}
& \textbf{33.0} & \textbf{55.3}
& \textbf{33.4} & \textbf{55.8}
\\
& \textsc{Rewrites}
& 23.5 & 44.2
& 24.0 & 44.7
\\
\bottomrule
\end{tabular}}
\caption{Overall performance of all model variants on \shortname{} development and test set}
\label{tab:main-results-table}
\end{table*}

\subsection{Evaluation metrics}
Following \newcite{choi&al18quac} and \newcite{reddy&al19coqa}, we use \textit{exact match (EM)} and \textit{F1} as evaluation metrics for \shortname{}.

To compute human and system performance in the presence of multiple gold annotations, we follow the evaluation process similar to \citet{choi&al18quac} and \citet{reddy&al19coqa}. Given $n$ human answers, human performance on the task is determined by considering each answer as prediction and other human answers as the reference set. This results in $n$ scores, which are averaged to give the final human performance score. The system prediction is also compared with $n$ distinct reference sets, each containing $n-1$ human answers, and then averaged. For \shortname{}, $n=4$ (the original answer and three additional annotations).
Note that human performance is not necessarily an upper bound for the task, as document retrieval can potentially be performed better by the systems.

\section{Results and Discussion}

We report the end-to-end performance of all systems in Table~\ref{tab:main-results-table}. For open-book models, we also look at the performance of its constituents (retriever and reader). Table~\ref{tab:retrieval-results-table} reports the retrieval performance and Table~\ref{tab:reader-results-table} reports the reading comprehension performance of the readers, given the gold passage. Based on these results, we answer the following research questions. \\

\noindent \textit{How do the models compare against humans for \shortname{}?}

\noindent We report model and human performance on development and test set in Table~\ref{tab:main-results-table}. Overall, model performance in all settings is significantly lower than the human performance. The best performing model (DPR Retriever + FiD using \textsc{AllHistory} question representation) achieves 33.4 points EM and 55.8 points F1
on the test set, which falls short of human performance by 6.9 points and 14.2 points respectively, indicating room for further improvement. \\

\noindent\textit{Which class of models perform better -- Closed book or Open book?}

\noindent GPT-3 is directly comparable to \textsc{AllHistory} variant of open-book models as it takes the entire conversation history as input. Apart from BM25 + DPR Reader, GPT-3 performs worse than all other \textsc{AllHistory} variants of open-book models. It achieves an F1 score of 31.8 on the test set, which is 24 points behind the best performing open-book model (DPR Retriever + FiD).
We observe that GPT-3 often hallucinates many answers, a phenomenon commonly observed in literature \cite{shuster2021retrieval}. \\

\noindent \textit{How does the performance of open-book models vary with various question representations?}

\noindent For all open-book models, we fine-tune on three different question representations (Section~\ref{sec:models}). 
From the results in Table~\ref{tab:main-results-table}, we observe that the \textsc{Original} representation is consistently worse than others for all models. This highlights the importance of encoding the conversational context for \shortname{}.
Between \textsc{AllHistory} and \textsc{Rewrites}, we observe that \textsc{AllHistory} performs better with dense retriever (DPR Retriever), whereas \textsc{Rewrites} performs better with sparse retriever (BM25).
To confirm that this performance difference in end-to-end systems stems from the retriever, we look at Top-20 and Top-100 retrieval accuracy of BM25 and DPR Retriever in Table~\ref{tab:retrieval-results-table}. Indeed, \textsc{AllHistory} representation performs better than \textsc{Rewrites} for DPR Retriever but worse for BM25. As DPR Retriever is trained on \shortname{}, it can probably learn how to select relevant information from the \textsc{AllHistory} representation, whereas for BM25,
the non-relevant keywords in the representation act as distractors. 
The better performance of DPR Retriever over BM25 indicates that \shortname{} requires learning task-specific dense semantic encoding for superior retrieval performance. \\

\begin{table}[t]
\resizebox{\linewidth}{!}{
\centering {\begin{tabular}{lcccccc}
\toprule       
Model & Question Rep & \multicolumn{2}{c}{Dev} & \multicolumn{2}{c}{Test} \\ 
\cmidrule(r){3-4} \cmidrule(r){5-6}

& & Top-20 & Top-100 & Top-20 & Top-100 \\

\midrule                         
& \textsc{Original}
& 5.2 & 9.1 & 6.0 & 10.1 \\
BM25 & \textsc{AllHistory}
& 23.1 & 36.8 & 22.5 & 35.6 \\
& \textsc{Rewrites}
& 32.5 & 49.2 & 33.0 & 47.4 \\
\midrule

& \textsc{Original}
& 9.9 & 16.5 & 10.0 & 15.3 \\
DPR Retriever & \textsc{AllHistory}
& \textbf{70.4} & \textbf{82.4} & \textbf{67.0} & \textbf{80.8} \\
& \textsc{Rewrites}
& 49.9 & 62.4 & 49.3 & 61.1 \\

\bottomrule
\end{tabular}}}
\caption{Retrieval performance of all model variants on \shortname{} development and test set}
\label{tab:retrieval-results-table}
\end{table}

\noindent \textit{How much are the readers constrained due to retrieved results?}

\noindent Table~\ref{tab:retrieval-results-table} shows retrieval results. In an end-to-end system, the reader take as input the retrieved passages, which may or may not contain the gold passage.
To get an estimate of reader performance independently from the retriever, we experiment with directly providing only the gold passage to the readers, instead of the retrieved ones. 
Table~\ref{tab:reader-results-table} shows the results.
This can be seen as an ``Ideal Retriever'' setting, where the retriever always retrieves the correct passage as the top one. 
Although we observe significant gains over end-to-end systems for all models across all variants, the best model (FiD with \textsc{AllHistory}) still falls short of human performance by 3.1 points EM and 5.9 points F1
on the test set. 
These experiments indicate that while passage retrieval is a significant bottleneck for the task, technical advancements are needed for the readers as well.

While it is plausible to assume that DPR Reader is restricted in its performance due to its extractive nature, we show that this is not the case.
We calculate the extractive upper bound for \shortname{} (reported in Table~\ref{tab:reader-results-table}), by selecting the span from the gold document with best F1 overlap with the ground truth answer. 
This bound is 47.3 points EM and 81.0 points F1,
which essentially represents the best that any extractive model can do on this task. DPR reader falls short of this upper bound by 19.2 points EM and 28.4 points F1.

\begin{table}[t]
\resizebox{\linewidth}{!}{
\centering {\begin{tabular}{lccccc}
\toprule       
Model & Question Rep & \multicolumn{2}{c}{Dev} & \multicolumn{2}{c}{Test} \\ 
\cmidrule(r){3-4} \cmidrule(r){5-6}

& & EM & F1 & EM & F1\\

\midrule      
Extractive Bound &
& \textbf{47.7}
& \textbf{81.1}
& \textbf{47.3}
& \textbf{81.0}
\\
\midrule
& \textsc{Original}
& 27.1
& 51.4
& 25.5
& 50.4
\\
DPR Reader & \textsc{AllHistory}
& 29.7
& 54.2
& 28.0
& 52.6
\\
& \textsc{Rewrites}
& 29.8
& 53.8
& 28.1
& 52.1
\\
\midrule

& \textsc{Original}
& 34.4
& 60.5
& 33.7
& 61.0
\\
FiD & \textsc{AllHistory}
& \textbf{38.3}
& \textbf{65.5}
& \textbf{37.2}
& \textbf{64.1}
\\
& \textsc{Rewrites}
& 34.5
& 61.9
& 35.3
& 62.8
\\

\bottomrule
\end{tabular}}}
\caption{Reader performance of all model variants on \shortname{} development and test set when provided with the gold passage}
\label{tab:reader-results-table}
\end{table}

\section{Conclusion}
We introduced \shortname{}, a novel open-domain conversational question answering dataset with topic switching. In this work, we described our data collection effort, analyzed its topic switching behaviour, and established strong neural baselines. The best performing model (DPR retriever + FiD) is 6.9 points EM and 14.2 points F1
below human performance, suggesting that advances in modeling are needed. We hope our dataset will be an important resource to enable more research on conversational agents that support topic switches in information-seeking scenarios. 

\section*{Acknowledgements}
We would like to thank \href{https://www.takt.tech}{TAKT's}
annotators and Jai Thirani (Chief Data Officer) for contributing to \shortname{}.
We are grateful for constructive and insightful feedback from the anonymous reviewers. 
This work is supported by the following grants: MSR-Mila Grant, NSERC Discovery Grant on \emph{Robust conversational models for accessing the world's knowledge} and the Facebook CIFAR AI Chair. 
Shehzaad is partly supported by the IBM PhD Fellowship. 
We thank Compute Canada for the computing resources.

\appendix

\section{Annotators Details}
\label{appendix_sec:data_collection}
Each conversation in \shortname{} is an interaction between two annotators -- a \textit{questioner} and an \textit{answerer}. The annotators were selected from TAKT's in-house workforce, based on their English language proficiency and trained for the role of both questioner and answerer. The annotators are provided with the following guidelines.

\noindent\textbf{Guidelines for the questioner:}
\begin{itemize}[noitemsep]
    \item The first question should be unambiguous and about the seed entity.
    \item The follow-up questions be contextualized and dependent on the conversation history whenever possible.
    \item Avoid using same words as in section titles of the document. E.g. if the section title is \textit{``Awards''}, a plausible question can be \textit{``What accolades did she receive for her work?''}.
    \item The conversation should involve multiple documents (topics).
\end{itemize}

\noindent\textbf{Guidelines for the answerer:}
\begin{itemize}[noitemsep]
    \item Based on the question, identify the relevant document and section.
    \item The answer should be based on the contents of the identified document.
    \item The rationale should be selected such that it justifies the answer.
    \item The answer should be a sub-string in rationale whenever possible. However, answers should be edited to fit the conversational context (adding \textit{yes}, \textit{no}), perform reasoning (e.g. counting) etc.
    \item Personal opinions should never be included.
\end{itemize}

\begin{figure}[ht!]
\footnotesize
\begin{tabular}{p{\columnwidth}}
\cmidrule(l{1em}r{1em}){1-1}
\texttt{Q$_1$}: \qstyle{when was the byzantine empire born what was it originally called?}\\
\texttt{A$_1$}: 5th century AD and was called Eastern Roman Empire, or Byzantium\\
\texttt{Topic}: \href{https://en.wikipedia.org/wiki/Byzantine_Empire}{Byzantine Empire}\\
\vspace{0em}
\texttt{Q$_2$}: \qstyle{and when did it fall?}\\
\texttt{A$_2$}: 1453\\
\texttt{Topic}: \href{https://en.wikipedia.org/wiki/Byzantine_Empire}{Byzantine Empire}\\
\vspace{0em}
\texttt{Q$_3$}: \qstyle{which battle or event marked the fall of this empire?}\\
\texttt{A$_3$}: A six-year-long civil war followed by attack from Sultan Mehmed's army\\
\texttt{Topic}: \href{https://en.wikipedia.org/wiki/Byzantine_Empire}{Byzantine Empire}\\
\vspace{0em}
\texttt{Q$_4$}: \qstyle{did he conquer other territories as well?}\\
\texttt{A$_4$}: Yes. Anatolia and in Southeast Europe as far west as Bosnia\\
\texttt{Topic}: \href{https://en.wikipedia.org/wiki/Mehmed_the_Conqueror}{Mehmed the Conqueror}\\
\vspace{0em}
\texttt{Q$_5$}: \qstyle{where is the first area located in present day terms?}\\
\texttt{A$_5$}: Turkey\\
\texttt{Topic}: \href{https://en.wikipedia.org/wiki/Anatolia}{Anatolia}\\
\vspace{0em}
\texttt{Q$_6$}: \qstyle{who were the oldest known inhabitants of this region?}\\
\texttt{A$_6$}: Mesopotamian-based Akkadian Empire\\
\texttt{Topic}: \href{https://en.wikipedia.org/wiki/Anatolia}{Anatolia}\\
\vspace{0em}
\texttt{Q$_7$}: \qstyle{what is the present day capital of the country?}\\
\texttt{A$_7$}: Ankara\\
\texttt{Topic}: \href{https://en.wikipedia.org/wiki/Turkey}{Turkey}\\
\vspace{0em}
\texttt{Q$_8$}: \qstyle{can you name some of the other major cities here?}\\
\texttt{A$_8$}: Istanbul\\
\texttt{Topic}: \href{https://en.wikipedia.org/wiki/Turkey}{Turkey}\\
\vspace{0em}
\texttt{Q$_{9}$}: \qstyle{were any of these cities associated with the first empire you were discussing?}\\
\texttt{A$_{9}$}: The Ottomans made the city of Ankara the capital first of the Anatolia Eyalet and then the Angora Vilayet\\
\texttt{Topic}: \href{https://en.wikipedia.org/wiki/Ankara}{Ankara}\\
\vspace{0em}
\texttt{Q$_{10}$}: \qstyle{what are some of the most famous landmarks in the second city?}\\
\texttt{A$_{10}$}: The obelisk, Valens Aqueduct, Column of Constantine, Church of the Saints Sergius and Bacchus\\
\texttt{Topic}: \href{https://en.wikipedia.org/wiki/Istanbul}{Istanbul}\\
\vspace{0em}
\texttt{Q$_{11}$}: \qstyle{who was the first monument you mentioned dedicated to?}\\
\texttt{A$_{11}$}: UNANSWERABLE\\
\texttt{Topic}:\\
\vspace{0em}
\texttt{Q$_{12}$}: \qstyle{and who was the third monument name after?}\\
\texttt{A$_{12}$}: Roman emperor Constantine the Great\\
\texttt{Topic}: \href{https://en.wikipedia.org/wiki/Column_of_Constantine}{Column of Constantine}\\
\vspace{0em}
\texttt{Q$_{12}$}: \qstyle{what is it made of?}\\
\texttt{A$_{12}$}: Porphyry and white marble\\
\texttt{Topic}: \href{https://en.wikipedia.org/wiki/Column_of_Constantine}{Column of Constantine}\\
\vspace{0em}
\texttt{Q$_{12}$}: \qstyle{how tall is it?}\\
\texttt{A$_{12}$}: The column's top is 34.8 m above the present-day ground level but the original height of the monument as a whole would have been nearly 50 m tall\\
\texttt{Topic}: \href{https://en.wikipedia.org/wiki/Column_of_Constantine}{Column of Constantine}\\
\cmidrule(l{1em}r{1em}){1-1}
\end{tabular}
\caption{A full conversation from \shortname{}.}
\label{fig:example_conversation_full}
\end{figure}

\newcommand{\rewritestyle}[1]{\textit{#1}}

\begin{figure}[ht!]
\footnotesize
\begin{tabular}{p{\columnwidth}}
\cmidrule(l{1em}r{1em}){1-1}
\texttt{Q$_1$}: \qstyle{harry potter and the chamber of secrets full book summary}\\
\texttt{Q$^\prime_1$}: \rewritestyle{harry potter and the chamber of secrets full book summary}\\
\texttt{A$_1$}: The plot follows Harry's second year at Hogwarts School
\ldots
investigate the attacks.\\
\vspace{0em}
\texttt{Q$_2$}: \qstyle{when was the book published?}\\
\texttt{Q$^\prime_2$}: \rewritestyle{when was harry potter and the chamber of secrets published?}\\
\texttt{A$_2$}: 2 June 1999\\
\vspace{0em}
. . . . .
\\
\vspace{0em}
\texttt{Q$_4$}: \qstyle{does it have any special connection with any other book in this series?}\\
\texttt{Q$^\prime_4$}: \rewritestyle{does the book have any special connection with any other book in the harry potter series?}\\
\texttt{A$_4$}: Yes, "Chamber of Secrets" has many links with the sixth book of the series, "Harry Potter and the Half-Blood Prince".\\
\vspace{0em}
\texttt{Q$_5$}: \qstyle{what is the storyline of that book?}\\
\texttt{Q$^\prime_5$}: \rewritestyle{what is the storyline of the sixth book of the series, harry potter and the half-blood prince?}\\
\texttt{A$_5$}: The novel explores the past of the boy wizard's nemesis, Lord Voldemort
\ldots
alongside his headmaster and mentor Albus Dumbledore.\\
\vspace{0em}
\texttt{Q$_6$}: \qstyle{what does the headmaster look like?}\\
\texttt{Q$^\prime_6$}: \rewritestyle{what does the headmaster of harry potter and the chamber of secrets look like?}\\
\texttt{A$_6$}: Albus Dumbledore is tall and thin
\ldots
twinkled with kindness and mischief.\\
. . . . .
\\
\vspace{0em}
\texttt{Q$_9$}: \qstyle{what is the real name of the boy wizard's nemesis mentioned above?}\\
\texttt{Q$^\prime_9$}: \rewritestyle{what is the real name of the boy wizard's nemesis mentioned above?}\\
\texttt{A$_9$}: Tom Marvolo Riddle\\
\vspace{0em}
\texttt{Q$_{10}$}: \qstyle{what are his magical strengths?}\\
\texttt{Q$^\prime_{10}$}: \rewritestyle{what are albus dumbledore's magical strengths?}\\
\texttt{A$_{10}$}: He is known as one of the greatest Legilimens in the world
\ldots
can read minds and shield his own from penetration.\\
\vspace{0em}
\texttt{Q$_{11}$}: \qstyle{is he related to any other family apart from the riddle one?}\\
\texttt{Q$^\prime_{11}$}: \rewritestyle{is tom marvolo riddle related to any other family apart from the riddle one?}\\
\texttt{A$_{11}$}: Yes, Gaunts\\
\vspace{0em}
\texttt{Q$_{12}$}: \qstyle{how is he related to them?}\\
\texttt{Q$^\prime_{12}$}: \rewritestyle{how is albus dumbledore related to the gaunts?}\\
\texttt{A$_{12}$}: They are the last known descendants of Salazar Slytherin.\\
\vspace{0em}
\texttt{Q$_{13}$}: \qstyle{in which city did the alchemist who worked with the headmaster live?}\\
\texttt{Q$^\prime_{13}$}: \rewritestyle{in which city did the alchemist who worked with the headmaster live?}\\
\texttt{A$_{13}$}: Paris\\

\cmidrule(l{1em}r{1em}){1-1}
\end{tabular}
\caption{An example of a conversation from \shortname{} along with rewrites from the $QR$ module. Few turns are excluded and some answers are shorted for brevity.}
\label{fig:example_conversation_full_harry_potter}
\end{figure}

\begin{figure*}[!htp]
\centering
\begin{subfigure}{\linewidth}
    \centering
    \includegraphics[height=175pt]{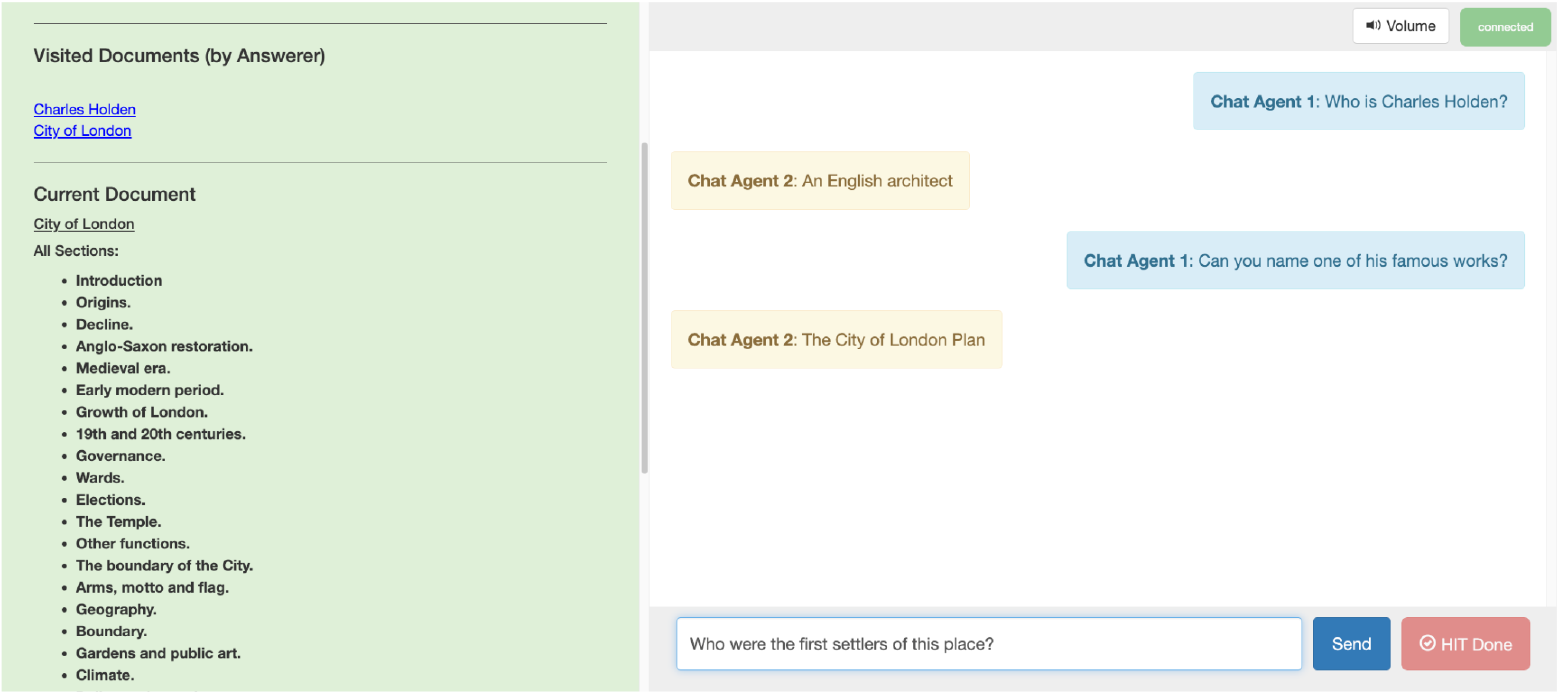}
    \caption{Questioner Interface}
\end{subfigure}
\hfill
\begin{subfigure}{\linewidth}
    \centering
    \includegraphics[height=157pt]{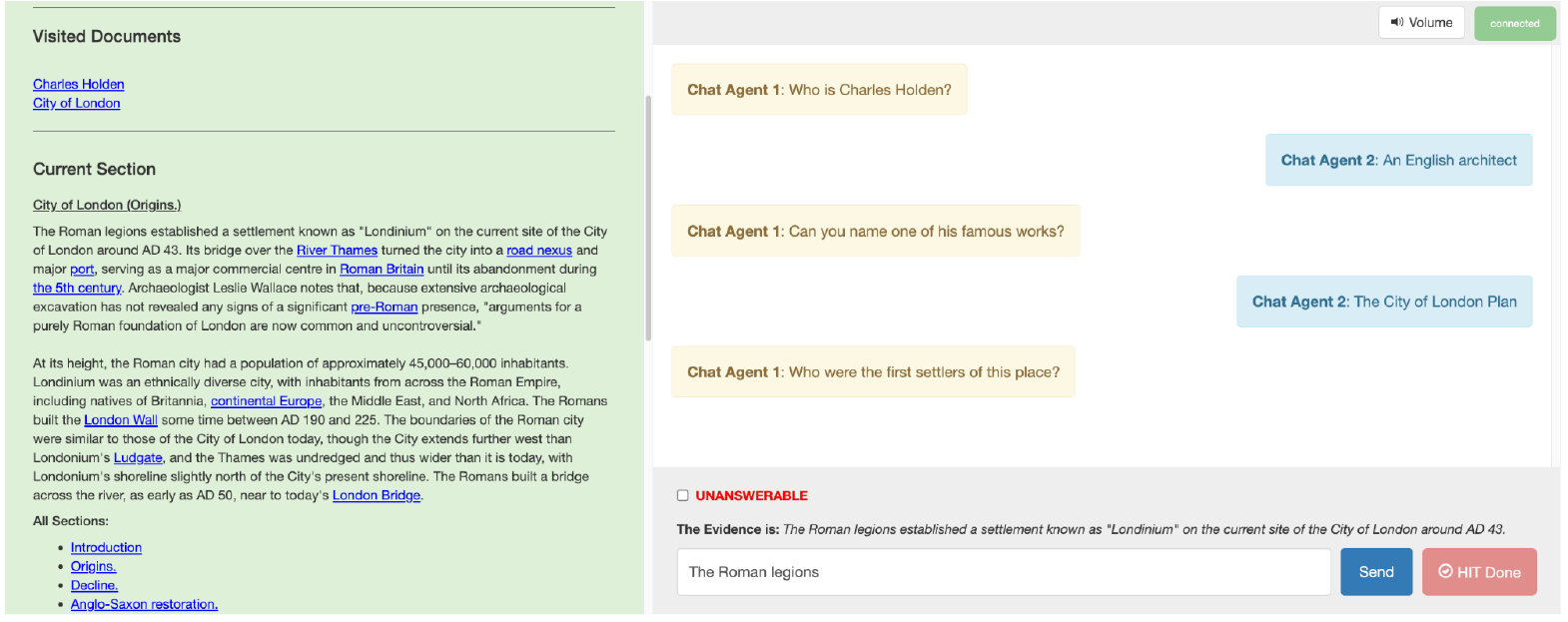}
    \caption{Answerer Interface}
\end{subfigure} 
\caption{Annotation interface for questioners and answerers}
\label{fig:interface-screenshot}
\end{figure*}

\noindent After providing the guidelines and a few examples, the initial annotated conversations were manually inspected by the authors. 
The workers that provided low-quality annotations during this inspection phase were disqualified.
The final workforce consisted of 15 workers, which provided annotations for the dataset over a period of two months. Random quality checks were performed by the authors and periodic feedback was given to the annotators throughout the data collection to maintain high quality of data.
Figure~\ref{fig:interface-screenshot} shows annotation interfaces for questioner and answerer.
Figure~\ref{fig:example_conversation_full} shows an example from the dataset.

We also implemented several real-time checks in the questioner's interface to encourage topic switching and use of co-reference, and to reduce the lexical overlap with the metadata of the document while forming the question.

\section{Query Rewriting}
\label{appendix_sec:query_rewriting}
A query-rewriting module $QR$, takes the current question and the conversation history as input $(q_1, a_1, \ldots, q_{n-1}, a_{n-1}, q_n)$ and provides a decontextualized rewritten question,  $q^\prime_n$, as the output. As we don't collect rewrites in \shortname{}, we rely on other datasets to train our $QR$ model. Two datasets that provide rewrites for information-seeking conversations are CANARD~\cite{elgohar&tal19canard} and QReCC~\cite{anantha&al21qrecc}. Due to its large-scale and diverse nature, we use QReCC to train our T5 model based $QR$ module.

To rewrite the $n^{th}$ question, the conversation history and the current question is given to model as $q_1$ \texttt{[SEP]} $a_1$ \texttt{[SEP]} $q_2$ \texttt{[SEP]} $a_2$ \texttt{[SEP]} $\ldots$ \texttt{[SEP]} $q_{n-1}$ \texttt{[SEP]} $a_{n-1}$ \texttt{[SEP]} $q_n$ .
We train this model on QReCC dataset. On the test split of QReCC, our model achieves a BLEU score of 62.74 points. We use this model to generate rewrites for \shortname{} in our experiments. Figure~\ref{fig:example_conversation_full_harry_potter} shows a conversation from the dataset along with rewrites from this T5-based $QR$ module.

We observe that while this $QR$ module can resolve simple coreferences (\texttt{Q$_2$} and \texttt{Q$_5$}), it struggles later in the conversation in the presence of multiple entities (\textit{he} is resolved to \textit{albus dumbledore} instead of \textit{tom marvolo riddle} in \texttt{Q$_{10}$} and \texttt{Q$_{12}$}). The $QR$ module also fails to perform reasoning required for correct rewrites, e.g - \textit{boy wizard's nemesis} is not rewritten to \textit{Lord Voldemort} in \texttt{Q$_9$}, even though this information is present in \texttt{A$_5$}).

\section{Hyperparameter Details}
\label{appendix_sec:hyperparam}
We use Lucene BM25 with $k1 = 0.9$ (term frequency scaling) and $b = 0.4$ (document length normalization). For both DPR and FiD, apart from the batch size, we use the hyperparameters suggested in their codebases. We use the maximum batch size that fits in the GPU cluster. DPR Retriever is trained on 4 40GB A100 GPUs, whereas DPR Reader and FiD are trained on 8 32GB V100 GPUs. We use \texttt{base} model size for all systems. Following original implementations, DPR Retriever is trained for 40 epochs, DPR Reader for 20 epochs, and FiD for 15000 gradient steps. The model checkpoint with best EM score on development set is selected as the final model.

\clearpage

\bibliography{custom}
\bibliographystyle{acl_natbib}

\end{document}